\documentclass[10pt,twocolumn,letterpaper]{article}

\usepackage[pagenumbers]{cvpr} 

\usepackage{graphicx}
\usepackage{amsmath}
\usepackage{amssymb}
\usepackage{booktabs}
\usepackage{enumitem}

\usepackage[pagebackref,breaklinks,colorlinks]{hyperref}

\usepackage[capitalize]{cleveref}
\crefname{section}{Sec.}{Secs.}
\Crefname{section}{Section}{Sections}
\Crefname{table}{Table}{Tables}
\crefname{table}{Tab.}{Tabs.}

\def\cvprPaperID{10621} 
\def\confName{CVPR}
\def\confYear{2023}

\begin{document}

\title{Multi-View Reconstruction using Signed Ray Distance Functions (SRDF)}

\author{
Pierre Zins$^{1,2}$~~~
Yuanlu Xu$^2$~~~
Edmond Boyer$^{1,3}$~~~
Stefanie Wuhrer$^1$~~~
Tony Tung$^2$~~~
\smallskip 
\vspace{-1mm}
\\
$^1$Univ. Grenoble Alpes, Inria, CNRS, Grenoble INP \thanks{Institute of Engineering Univ. Grenoble Alpes}, LJK, 38000 Grenoble, France
\\
$^2$Meta Reality Labs, Sausalito, USA
\\
$^3$Meta Reality Labs, Zurich, Switzerland
\\
\small{\textit{ name.surname@inria.fr, merayxu@gmail.com, tony.tung@fb.com}}
}

\maketitle

\begin{abstract}
In this paper, we investigate a new optimization framework for  multi-view 3D shape reconstructions. Recent differentiable rendering approaches  have provided breakthrough performances with implicit shape representations though they can still lack precision in the estimated geometries. On the other hand multi-view stereo methods can yield pixel wise geometric accuracy with local depth predictions along viewing rays. Our approach bridges the gap between the two strategies with  a novel  volumetric shape representation that is implicit but parameterized with pixel depths to better materialize the shape surface with consistent signed distances along viewing rays. The approach retains pixel-accuracy while benefiting from volumetric integration in the optimization. To this aim, depths are optimized by evaluating, at each 3D location within the volumetric discretization, the agreement between the depth prediction consistency and the photometric consistency for the corresponding pixels. The optimization is agnostic to the associated photo-consistency  term which can vary from a median-based baseline to more elaborate criteria~\eg learned functions. Our experiments demonstrate the benefit of the volumetric integration with depth predictions. They also show that our approach outperforms existing approaches over standard 3D benchmarks with better geometry estimations.

\end{abstract}

\section{Introduction}
Reconstructing 3D shape geometries from  2D image observations has been a core issue in computer vision
for decades. Applications are numerous and range from robotics to augmented reality and human digitization, among others. 
When images are available in sufficient numbers, multi-view stereo (MVS) is a powerful
strategy that has emerged in the late 90s (see~\cite{seitz2006}). In this strategy, 3D geometric models are built by  searching for surface locations in  3D where 2D image observations concur, a property called photo-consistency. This observation consistency strategy has been later challenged  by  approaches in the field that seek instead for observation fidelity using differentiable rendering. Given a shape model that includes appearance information, rendered images can be compared to observed images and the model can thus be optimized. Differentiable rendering adapts to several shape representations including  point clouds, meshes  and, more recently, implicit shape representations. The latter can account for occupancy, distance functions or densities, which are estimated either directly over discrete grids or through continuous MLP network functions. Associated to differentiable rendering these implicit representations have provided  state-of-the-art approaches to recover both the geometry and the  appearance of 3D shapes from 2D images. \\

\begin{figure}[ptb]
\begin{center}
\includegraphics[width=\linewidth]{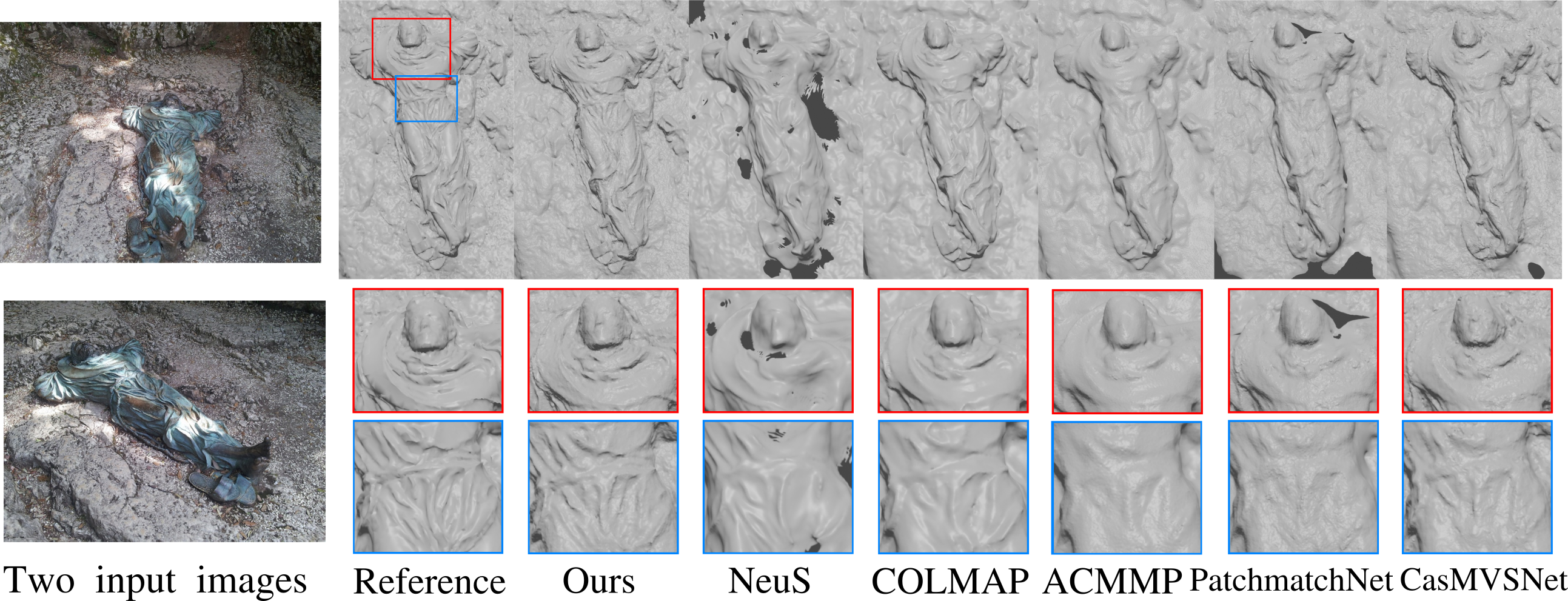}
\end{center}
\vspace{-3mm}
\caption{Reconstructions with various methods using $14$ images of a model from BlendedMVS~\cite{yao2020blendedmvs}.}
\label{fig:teaser}
\vspace{-4mm}
\end{figure}

With the objective to improve the precision of the reconstructed geometric models and their computational costs, we investigate an approach that takes inspiration from differentiable rendering methods while retaining beneficial aspects of MVS strategies. Following volumetric methods we use a volumetric signed ray distance representation which we parameterize with depths along viewing rays, a representation we call the Signed Ray Distance Function or SRDF. This representation makes the shape surface explicit with depths while keeping the benefit of better distributed gradients with a volumetric discretization. To optimize this shape representation we introduce an unsupervised differentiable volumetric criterion that, in contrast to differentiable rendering approaches, does not require color estimation. Instead, the criterion considers volumetric 3D samples and evaluates  whether the signed distances along rays agree at a sample when it is photo-consistent and disagree otherwise. While being volumetric our proposed approach shares the following MVS benefits: 
\begin{itemize}[noitemsep,nolistsep]
    \item[i)] No expensive ray tracing in addition to color decisions is required; 
    \item[ii)] The proposed approach is pixel-wise accurate by construction; 
    \item[iii)] The optimization can be performed over groups of cameras defined with visibility considerations.  The latter enables parallelism between groups while still enforcing consistency over depth maps. 
\end{itemize}
In addition, the volumetric scheme provides a testbed to compare  different photo-consistency  priors  in a consistent way with space  discretizations that do not depend on the estimated surface.

To evaluate the approach, we conducted experiments on real data from DTU Robot Image Data Sets~\cite{jensen2014large}, BlendedMVS~\cite{yao2020blendedmvs} and on synthetic data from Renderpeople~\cite{renderpeople} as well as on real human capture data. Ablation tests demonstrate the respective contributions of the SRDF parametrization and the volumetric integration in the shape reconstruction process. Comparisons with both MVS and Differential Rendering methods also show that our method consistently outperforms state of the art both quantitatively and qualitatively with better geometric details.

\section{Related Work}

\subsection{Multi-view Stereo}
Reconstructing 3D shapes from multiple images is a long-standing problem in computer vision. Traditional MVS approaches can be split into two categories. Seminal methods~\cite{de1999poxels, broadhurst2001probabilistic, seitz1999photorealistic, kutulakos2000theory} use a voxel grid representation and try to estimate occupancies and colors. While efficient their reconstruction precision is inherently limited by  the  memory requirement of the 3D grid when increasing resolution. On the other hand depth-based methods~\cite{ furukawa2009accurate, agrawal2001probabilistic, campbell2008using, galliani2016gipuma, schonberger2016pixelwise, xu2019multi} have been proposed that usually try to match image features from several views to estimate depths. Additional post-processing fusion and meshing steps~\cite{curless1996volumetric, merrell2007real, kazhdan2013screened, kazhdan2006poisson} are required to recover a surface from the multi-view depth maps. Despite the usually complex pipeline, multi-view depth map estimation offers the advantage to give access to pixel-accuracy, a strong feature  for the reconstruction quality which has made this strategy common in MVS approaches. We also consider multi-view depth maps that are however included into a volumetric framework, with signed distances, with the aim to improve global consistency.
With the advances in deep learning, several methods propose to learn some parts of the MVS pipeline such as the image feature matching~\cite{hartmann2017learned, zagoruyko2015learning, leroy2018shape} or the depth map fusion~\cite{donne2019learning, riegler2017octnetfusion}. Others even propose to learn the full pipeline in an end-to-end manner~\cite{huang2018deepmvs, yao2018mvsnet, yao2019recurrent, zagoruyko2015learning, sun2021neuralrecon, murez2020atlas, rich20213dvnet, wang2021patchmatchnet, gu2020cascade}. These learning based methods offer fast inference and exhibit interesting generalization abilities. Optimization based methods can be seen as alternative or complementary solutions with better precision and generalization abilities, as shown in our experiment in sec.~\ref{sec:finetuning}.

\subsection{Differentiable Rendering}
Another line of works has explored differentiable rendering approaches. Many of these works are used for novel view rendering  applications however several  also consider 3D shape geometry reconstruction, often as part of the  new image generation process. They build on a rendering that is differentiable and henceforth enable shape model optimization by differentiating the discrepancy between generated and observed images.  These methods were originally applied to various shape representations including meshes~\cite{liu2019soft, kato2018neural, henderson2018learning, passalis2022opendr}, volumetric grids~\cite{tulsiani2017multi, gadelha20173d, jimenez2016unsupervised, zhu2018visual, nguyen2018rendernet} or even point clouds~\cite{jiang2018gal, insafutdinov2018unsupervised, navaneet2019capnet}. In association with deep learning, new neural implicit representations have also emerged. Their continuous nature and light memory requirements are attractive and they have been successfully applied to different tasks: 3D reconstruction~\cite{saito2019pifu, saito2020pifuhd, xu2019disn, mescheder2019occupancy, peng2020convolutional, park2019deepsdf} or geometry and appearance representations~\cite{genova2019learning, takikawa2021neural, michalkiewicz2019implicit, oechsle2019texture, sitzmann2019scene}. Most of these methods solve for 3D shape inference and require 3D supervision, however recent works combine implicit representations with differentiable renderering and solve therefore for shape optimization with 2D image supervision. They roughly belong to two categories depending on the shape representation they consider. 

\begin{figure*}[pt]
\begin{center}
\includegraphics[width=0.65\linewidth]{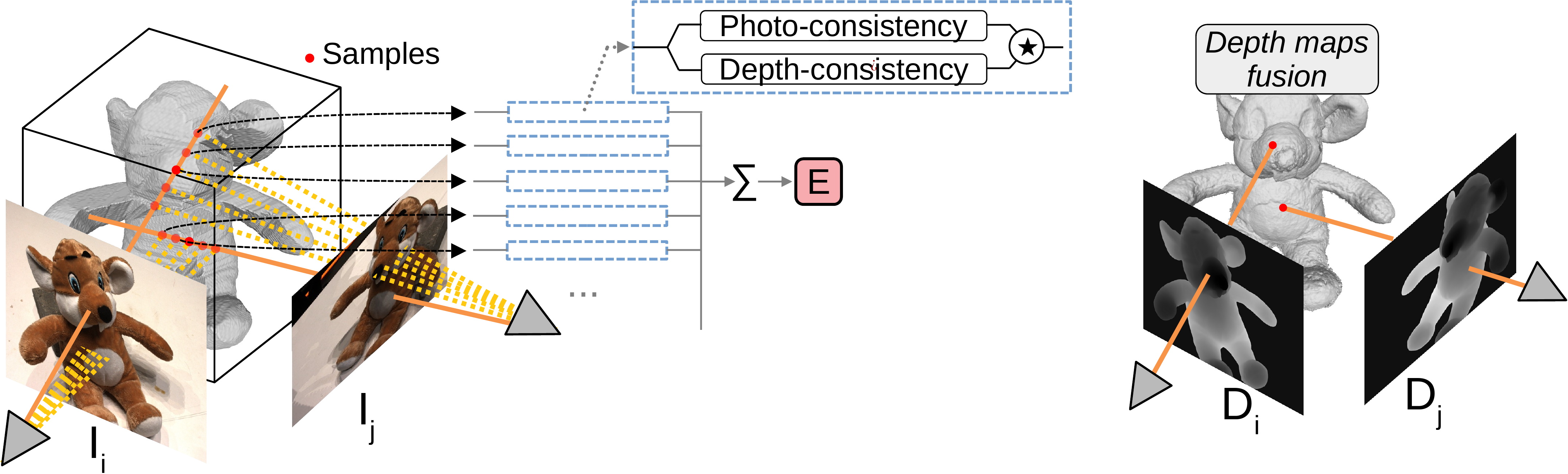}
\end{center}
\vspace{-3mm}
\caption{Our method overview. Left: Given multiple RGB images and an initial coarse reconstruction, our method optimizes depth maps using a volumetric shape energy $E$ that is evaluated at samples along viewing lines. Right: The optimized depth maps are further fused into a surface model.}
\label{fig:overview}
\vspace{-3mm}
\end{figure*}

\textbf{Volume-based methods} use a volumetric renderer with 3D samples and estimate for each sample a density as well as a color conditioned on a viewing direction. Colors and densities of the samples are integrated along viewing rays to obtain image colors that can be compared with the observed pixel colors. The pioneer work of Mildenhall \etal~\cite{mildenhall2020nerf} opened up this research area with impressive results on novel view synthesis. The quality of the associated geometry as encoded with densities is however not perfect and often noisy. Several works have followed that target generalization to new data~\cite{yu2021pixelnerf, jang2021codenerf}, dynamic scenes~\cite{pumarola2021d, li2021neural} or propose new formulations based on Signed Distance Functions to improve the estimated geometry~\cite{wang2021neus, yariv2021volume}. \cite{darmon2022improving} also proposes a finetuning strategy based on image warpings to take advantage of high-frequency texture. A main limitation of methods based on neural volumetric rendering lies in the optimization time complexity which often plagues the shape modeling process. To address this limitation various strategies have been investigated: a divide-and-conquer strategy~\cite{reiser2021kilonerf}, more efficient sampling~\cite{arandjelovic2021nerf}, traditional volumetric representations to directly optimize densities and colors inside octrees~\cite{yu2021plenoctrees}, or voxel grids~\cite{yu2021plenoxels} and  efficient multi-resolution hash tables~\cite{muller2022instant}. See~\cite{dellaert2020neural} for more details. These strategies dramatically decrease the optimization time while maintaining very good results for novel view rendering. However the estimated geometry can still lack precision as the methods are not primarily intended to perform surface reconstruction.

\textbf{Surface-based methods}~\cite{liu2020dist, yariv2020multiview, DBLP:journals/corr/abs-1912-07372, kellnhofer2021neural} address this issue with a surface renderer that estimate 3D locations where viewing rays enter the surface and their colors. By making the shape surface explicit, these approaches obtain usually better geometries. Nevertheless, they are more prone to local minima during the optimization since gradients are only computed near the estimated surface as opposed to volumetric strategies. An interesting hybrid approach~\cite{oechsle2021unisurf}, proposes to combine the advantages of both volumetric and surface rendering and shows good surface reconstructions.

\section{Method}

Our method takes as input $N$ calibrated color images $\mathcal{I=}I_{j \in [1,N]}$ and assumes $N$ initial associated depth maps $\mathcal{D=}D_{j \in [1,N]}$ that can be  obtained using an initial coarse reconstruction, with for instance pre-segmented image silhouettes as in Figure~\ref{fig:overview}. It optimizes depth values along pixel viewing lines by considering a photo-consistency criterion that is evaluated in 3D over an implicit volumetric shape representation. Final shape surfaces are thus obtained by fusing depth maps, as in \eg~\cite{curless1996volumetric,galliani2016gipuma}. The main features of the method are: 
\begin{itemize}
\item Shape representation (Sec~\ref{sec:SRDF}): depth maps determine the signed distances, along pixel viewing rays, that define our volumetric shape representation with the SRDF. Parameterizing with depths offers several advantages: it better accounts for the geometric context by materializing the shape surface; it enables pixel accuracy regardless of the image resolution; it allows for coarse to fine strategies as well as parallelization with groups of views.   
\item Energy function (Sec~\ref{sec:differentiable_volumetric_cost}): our shape energy function is evaluated at sample locations along viewing lines and involves multiple depth maps simultaneously, therefore enforcing spatial consistency. It focuses on the geometry and avoids potential ambiguous estimation of the  appearance.
\item Photometric prior (Sec~\ref{sec:photometric_prior}): The photo-consistency hypothesis evaluated by the energy function along a viewing line can be diverse. We propose a criterion that is learned over ground truth 3D data such as DTU~\cite{jensen2014large}. We also experiment a baseline unsupervised criterion that builds on the median color. 
\end{itemize}

\subsection{Signed Ray Distance Function}
\label{sec:SRDF}

Our shape representation is a volumetric signed distance function parameterized by depths along viewing rays. This is inspired by signed distance functions (SDF) and shares some similarities with more recent works on signed directional distance functions (SDDF)~\cite{zobeidi2021deep}. Unlike traditional surface-based representations such a function is differentiable at any point in the 3D observation volume.

\begin{figure}[t]
\begin{center}
\includegraphics[width=0.55\linewidth]{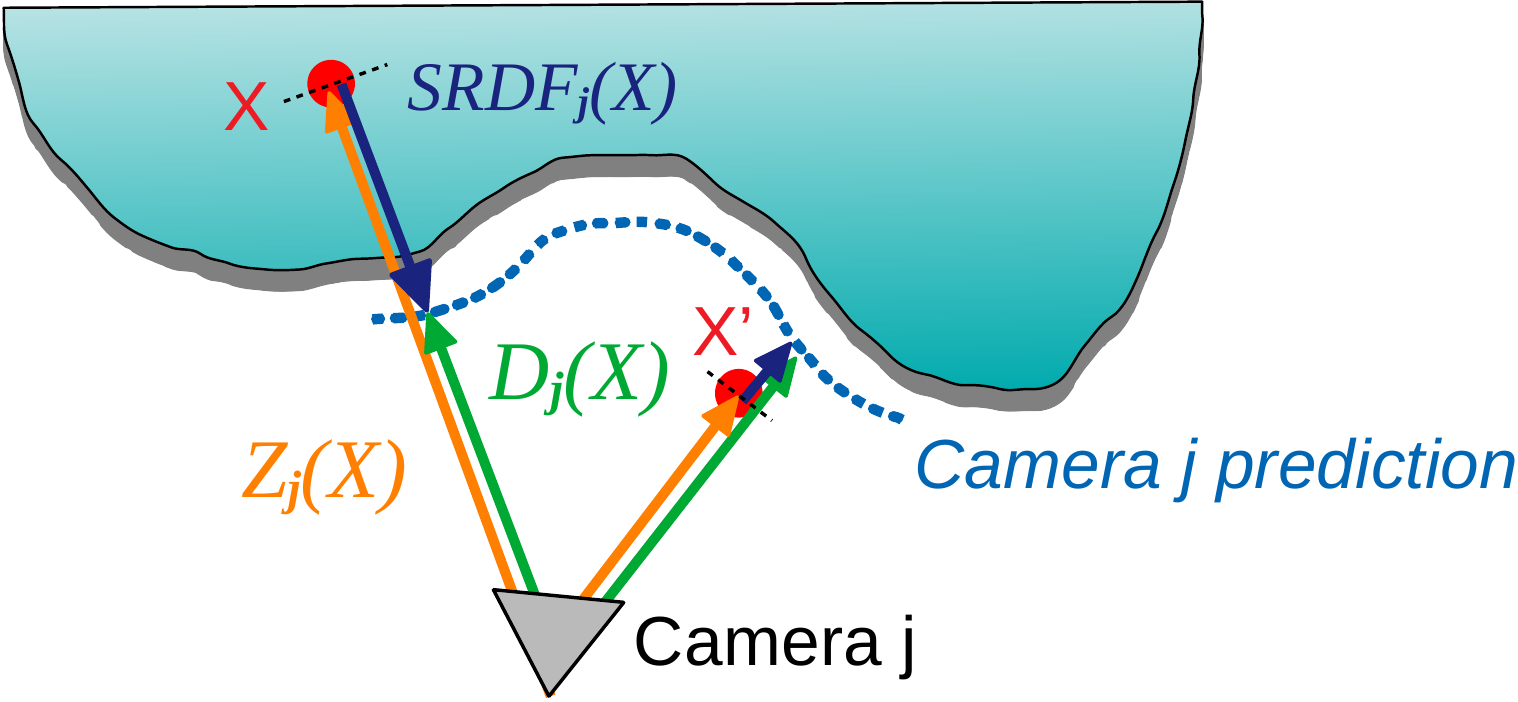}
\end{center}
\vspace{-3mm}
\caption{For any 3D point $X$, its ray signed distance $SRDF_j(X)= D_j(X) - Z_j(X)$ with respect to camera $j$ is the signed shortest distance from $X$ to the surface, as predicted by camera $j$  along the corresponding viewing line.}
\label{fig:raysdf}
\vspace{-3mm}
\end{figure}

Instead of considering the shortest distances along any direction as in standard SDF, or in a fixed direction as in SDDF~\cite{zobeidi2021deep}, we define, for a given 3D point $X$, its $N$  signed distances  with respect to cameras $j \in [1,N]$ as the signed distances of $X$ to its nearest neighbor on the surface as predicted by camera $j$ along the viewing ray passing through $X$. We denote the distance for $X$ and camera $j$ by the \emph{Signed Ray Distance Function (SRDF)}, as illustrated in Figure~\ref{fig:raysdf}:
\begin{small}
\begin{equation}
SRDF(X,D_j) =SRDF_j(X)  = D_j(X) - Z_j(X),
\end{equation}
\end{small}
where $D_j(X)$ is the depth in depth map $D_j$ at the the projection of $X$  and $Z_j(X)$ the distance from $X$ to camera $j$.

\subsection{Volumetric Shape Energy}
\label{sec:differentiable_volumetric_cost}

\begin{figure}[t]
\begin{center}
\includegraphics[width=0.9\linewidth]{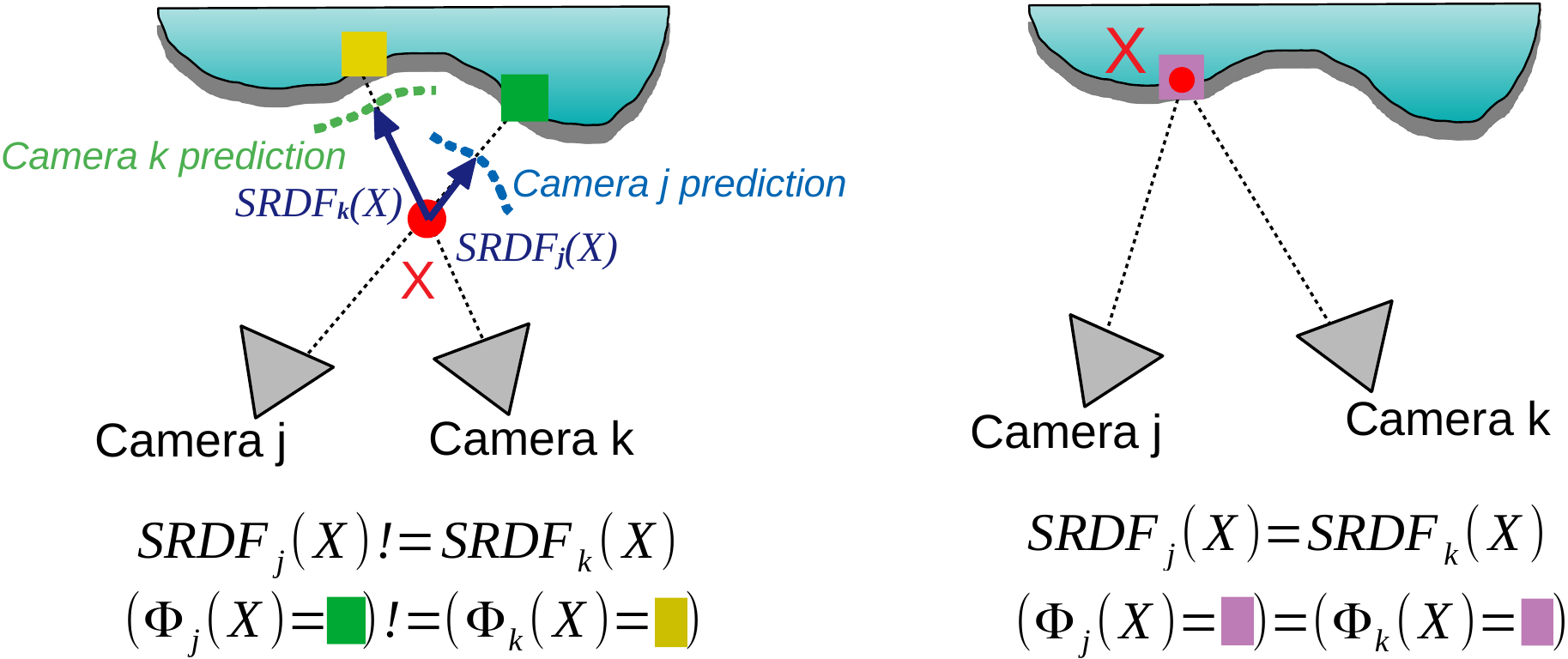}
\end{center}
\vspace{-3mm}
\caption{Inconsistency (left) and consistency (right) of the ray signed distances $SRDF_{j,k}(X)$ and of the photometric information $\Phi_{j,k}(X)$ at $X$ with respect to cameras $j$ and $k$.}
\label{fig:raysdf2}
\vspace{-3mm}
\end{figure}

\begin{figure}[!]
\begin{center}
\includegraphics[width=0.8\linewidth]{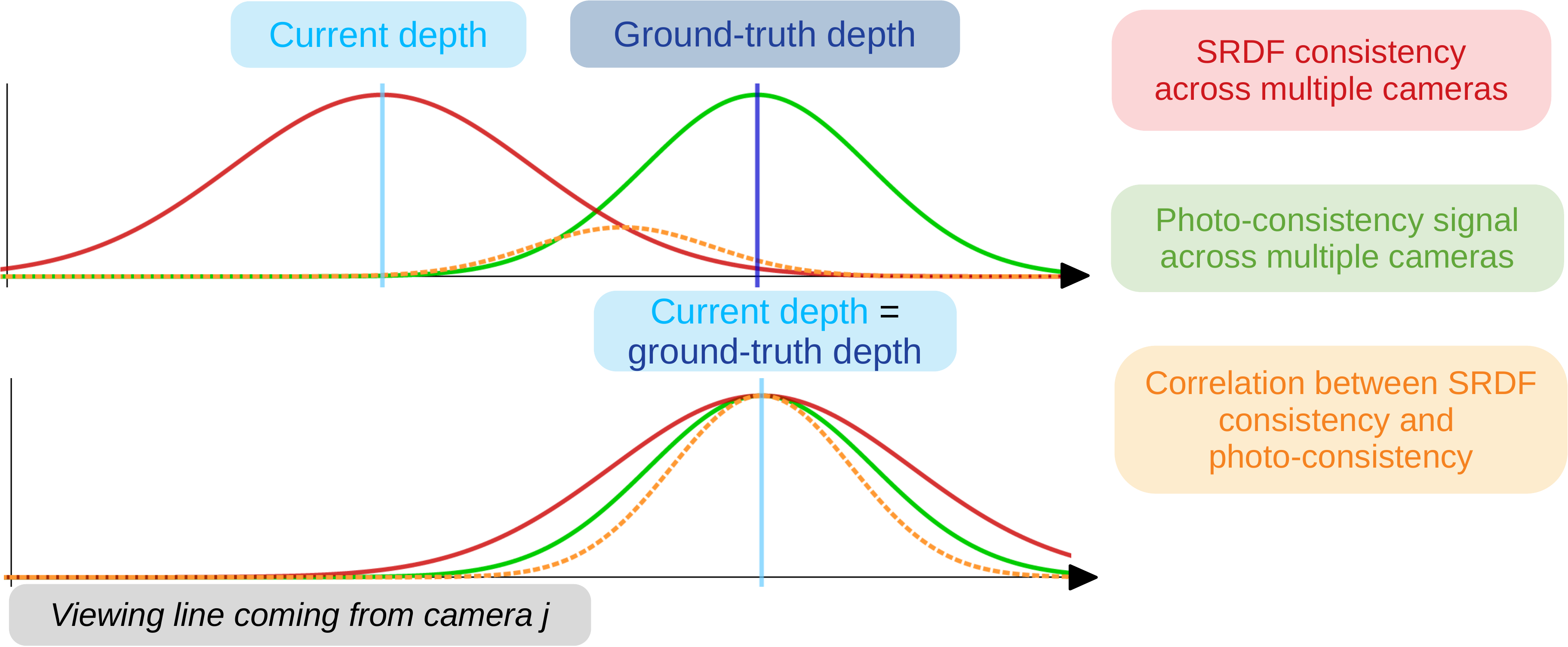}
\end{center}
\vspace{-3mm}
\caption{The SRDF consistency (red) and photo-consistency signals (green) along a viewing line. Their cross correlation will be maximal when the current predicted depth aligns with the ground truth depth.}
\label{fig:cost}
\vspace{-3mm}
\end{figure}

The intuition behind our volumetric energy function is that photometric observations  across different views should be consistent on the surface and not elsewhere. Importantly such a behavior
should be shared by the SRDF predictions across views  that should also consistently identify  zero distances for points  on the surface and non consistent distances elsewhere.   Given this principle, illustrated in Figure~\ref{fig:raysdf2}, a computational strategy is to look at the correlation between these 2 signals, the observed photo-consistency and the predicted SRDF consistencies, and to try to maximise it at 3D sample locations $\{X\}$ in the observation space (see Figure~\ref{fig:cost}). To this purpose we introduce the following consistency energy  function:

\begin{small}
\vspace{-3mm}
\begin{equation}
\label{eq_cost_1}
E(\{X\},\mathcal{D},\mathcal{I}) = \sum_{X} C_{SRDF}(X,\mathcal{D}) ~~C_{\Phi}(X,\mathcal{I}),
\end{equation}
\end{small}
where $\{X\}$ are the 3D sample locations, $C_{SRDF}(X,\mathcal{D})$ and $C_{\Phi}(X,\mathcal{I})$ represent   measurements of consistency among  the predicted SRDFs values $SRDF_{j \in [1,N]}(X)$ and among the observed photometric observations $\Phi_{j \in [1,N]}(X)$, respectively, at location $X$. Both are functions that return values between $0$ and $1$ that characterize consistency at  $X$. We detail below the SRDF consistency measure $C_{SRDF}(X)$. The photo-consistency measure $C_{\Phi}(X)$ is discussed in Section~\ref{sec:photometric_prior}. The above energy $E$ is differentiable  with respect to the predicted depths values $\mathcal{D}$ and computed in practice at several sample locations along each viewing ray of each camera, which enforces SRDFs to consistently predict surface points over all cameras.\\

\textbf{SRDF consistency}
From the observation that SRDF consistency is only achieved when $X$ is on the surface,~\ie when $SRDF_j(X) = 0$ for all non occluded cameras $j$, we define:
\begin{small}
\begin{equation}
\label{eq:SRDF_consistency}
    C_{SRDF}(X) = \prod_{j=1}^n \Big( exp\Big(-\frac{SRDF_j(X)^2}{\sigma_d}\Big) + \Gamma_{SRDF} \Big),
\end{equation}
\end{small}
where the ray signed distances are transformed into probabilities using an exponential which is maximal when $SRDF_j(X) = 0$. $\Gamma_{SRDF}$ is a constant that  prevents the product over all cameras to cancel out in case of inconsistencies caused by camera occlusions. It can be interpreted as the probability of the $SRDF_j$ value at $X$ knowing camera $j$ is occluded, which can be set as constant for all values. $\sigma_d$ is a hyper parameter that controls how fast  probabilities decrease with distances to the surface. It should be noted here  that the above energy term $C_{SDRF}$ is a product over views at a 3D point $X$ and not a sum, hence gradients \wrt depth values are not independent at $X$, which forces distances to become consistent across views, as shown in the ablation test provided in the supplemental.

\subsection{Photometric Consistency}
\label{sec:photometric_prior}

Our model is agnostic to the photo-consistency measure $C_{\Phi}(X)$  that is chosen.  In practice we have considered $2$ instances of $C_{\Phi}(X)$: A baseline version that relies on the traditional Lambertian prior assumption for the observed surface and a learned   version that can be trained with  ground truth 3D data. 

\textbf{Baseline Prior} assumes a Lambertian surface and therefore similar photometric observations for points on the observed surface for all non-occluded viewpoints. While ignoring non diffuse surface reflections the assumption has been widely used in image based 3D modelling, especially by MVS strategies. The associated consistency measure we propose accounts for the distance to the median observed value. Under the Lambertian assumption all observed appearances from non-occluded viewpoints should be equal. Assuming further that occluded viewpoints are fewer we define the photo-consistency as: 

\begin{small}
\vspace{-3mm}
\begin{equation}
\label{eq:baseline_prior}
    C_\Phi(X) = \\
    \prod_{j=1}^n \Big( exp\Big(-\frac{\lVert \Phi_j(X) -  \widetilde{\Phi}(X,\mathcal{I}))\rVert^2}{\sigma_c}\Big) + \Gamma_{\Phi} \Big),
\end{equation}
 \end{small}
where $\Phi_j(X)$ is photometric observation of $X$ in image $j$, typically a RGB color, and $\widetilde{\Phi}(X,\mathcal{I}))$ is the median  value of the observations at $X$  over all images.  Similarly to equation~\ref{eq:SRDF_consistency}, $\Gamma_{\Phi}$ is a constant that prevents the product over all cameras to cancel out in case of occlusion and $\sigma_c$  is a hyper-parameter. As shown in Section~\ref{sec:experiments} this baseline photometric prior yields state-of-the-art results on synthetic 3D data for which the Lambertian assumption holds but is less successful on real data.

\begin{figure}[t]
\begin{center}
\includegraphics[width=0.9\linewidth]{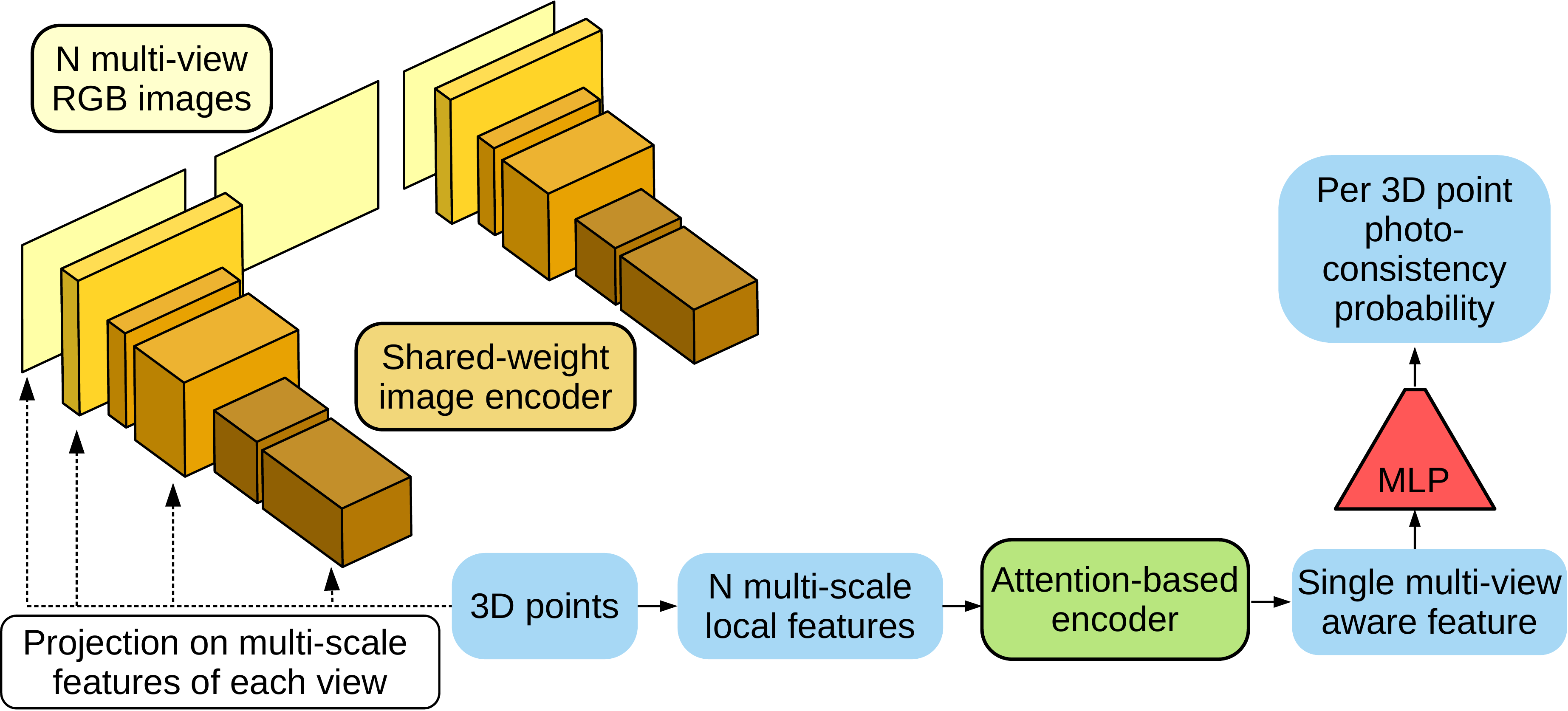}
\end{center}
\vspace{-3mm}
\caption{Proposed architecture to learn photo-consistency.}
\label{fig:archi}
\vspace{-3mm}
\end{figure}

\textbf{Learned Prior}. In order to better handle real images that are noisy and for which the Lambertian assumption is partially or not satisfied, we have experimented a more elaborated photo-consistency measure with a data driven approach. Inspired by previous works \cite{leroy2018shape, hartmann2017learned}, we cast the problem as a classification task between points that are photo-consistent across multiple views and points that are not and train a network for that purpose.

As described in Figure~\ref{fig:archi}, the network architecture tries to match the local appearance of a 3D point in different views and outputs a photo-consistency score between $0$ and $1$. This module is independent of the number of cameras and provides very good results on real data and generalization abilities as demonstrated in Section~\ref{sec:experiments}.
Please refer to the supplementary materials for more details about the architecture.

\section{Implementation}
\label{sec:implementation}

\subsection{Optimization Pipeline}
To allow for efficient processing, we define $G$ groups of cameras, which can be optimized in parallel. Since our approach optimizes geometry based on appearance matching, it is preferable to minimize occlusions. For this reason, we heuristically choose to gather cameras that are close to each other.

For the depth maps associated to a camera group, at each epoch, we iterate over all rays $r_j^i$ corresponding to foreground pixels $i$ of cameras $j$, as defined by pre-segmented silhouettes, and sample points along $r_j^i$ around the current depth estimation $d_j^i$. This sampling is parameterized by two parameters: an offset $o$ that defines an interval for the sampling around the current depth $[d_j^i - o; d_j^i + o]$, and the density of the sampling which represents the number of points that we sample uniformly in that interval.
Ideally, the real depth $\hat{d_j^i}$ should be contained inside the interval $ [d_j^i - o; d_j^i + o]$, otherwise it is difficult for the appearance  to guide the geometry optimization. From that observation, we define a coarse-to-fine strategy for the sampling. With the aim to capture the ground truth depth in the interval, we start with a large interval that is gradually reduced. The sampling density can be adjusted in the same way but decreasing the size of the sampling interval already indirectly increases its density, so in practice we keep the sampling density constant.

Our shape energy, described in subsection~\ref{sec:differentiable_volumetric_cost}, is computed over all the samples from all the rays of each camera. The gradients are computed using Pytorch autodiff~\cite{paszke2017automatic} and back-propagated to update depth maps.

\subsection{Photo-consistency Network}
\label{sec:photo_consistency_network}

To train the photo-consistency network, we use the DTU Robot Image Data Sets~\cite{jensen2014large} composed of $124$ scans of objects. For each scan, there are $49$ or $64$ images under $8$ different illuminations settings, camera calibration and ground truth point cloud obtained from structured light. We select $15$ test objects and remove all the scans that contain these objects from the training set which results in $79$ training scans. Next, we reconstruct a surface from the ground truth point cloud using the Screened Poisson algorithm~\cite{kazhdan2013screened} and surface trimming of $9.5$. From the reconstructed meshes we render ground truth depth maps and use them to sample points on the surface (positive samples) and points that are either in front or behind the surface (negative samples). We make sure to keep a balanced sampling strategy with an equal number of positive and negative samples.

To encourage the network to remain invariant to the number of cameras, at each training iteration, we randomly select a subset of $K$ cameras from the total $N$ cameras. Matching appearances between cameras too far from each other leads to inconsistencies as a result of the potentially high number of occlusions. To remedy this, we create the camera groups using a soft nearest neighbour approach. We randomly select a first camera, compute its $K'$ nearest neighbours cameras with $K < K' < N$, and randomly select $K-1$ cameras from them. In practice, $N = 49$ or $64$ and we choose $K \in [4,10]$ and $K' = min(2K, 15)'$.

\section{Experimental Results}
\label{sec:experiments}

To assess our method we conduct an evaluation on multi-view 3D shape reconstruction. First, we introduce the existing methods that we consider as our baseline. Then, we present the datasets as well as the evaluation metrics. We provide quantitative and qualitative comparisons against the current state of the art on real images using our learned prior for photo-consistency. Then, we also show that our method combined with a baseline prior for photo-consistency provides good reconstruction results under the Lambertian surface assumption. Finally, we demonstrate better generalization abilities of our method compared to deep MVS inference-based methods and that the latter can serve as an initialization. The values of our hyper parameters for each experiment are available in the supplementary. \textit{Code will be publicly available once this paper gets accepted}.

\subsection{Datasets and Metrics}
\label{sec:dataset}
To evaluate our method on real multi-view images with complex lighting, we use the 15 test objects from the DTU Data Sets~\cite{jensen2014large} and BlendedMVS~\cite{yao2020blendedmvs}. Note again that BlendedMVS is not used to train our learned photo-consistency prior. For DTU, the corresponding background masks are provided by~\cite{yariv2020multiview}. To test our method with the baseline prior for photo-consistency, we render multi-view images from Renderpeople~\cite{renderpeople} meshes. This dataset provides highly detailed meshes obtained from 3D scans of dressed humans and corrected by artists. We render $19$ high-resolution images (2048x2048) that mostly show the frontal part of the human.
For the quantitative evaluation with DTU we use a Python implementation~\cite{dtu_eval} of the official evaluation procedure of DTU. The accuracy and completeness metrics, with the Chamfer distances in $mm$, are computed~\wrt ground truth point clouds obtained from structured light.
Finally, to evaluate generalization to novel data, we also experiment with images from a large scale hemispherical multi-view setup with $65$ cameras of various focal lengths.

\begin{figure*}[t]
\begin{center}
\includegraphics[width=0.9\linewidth]{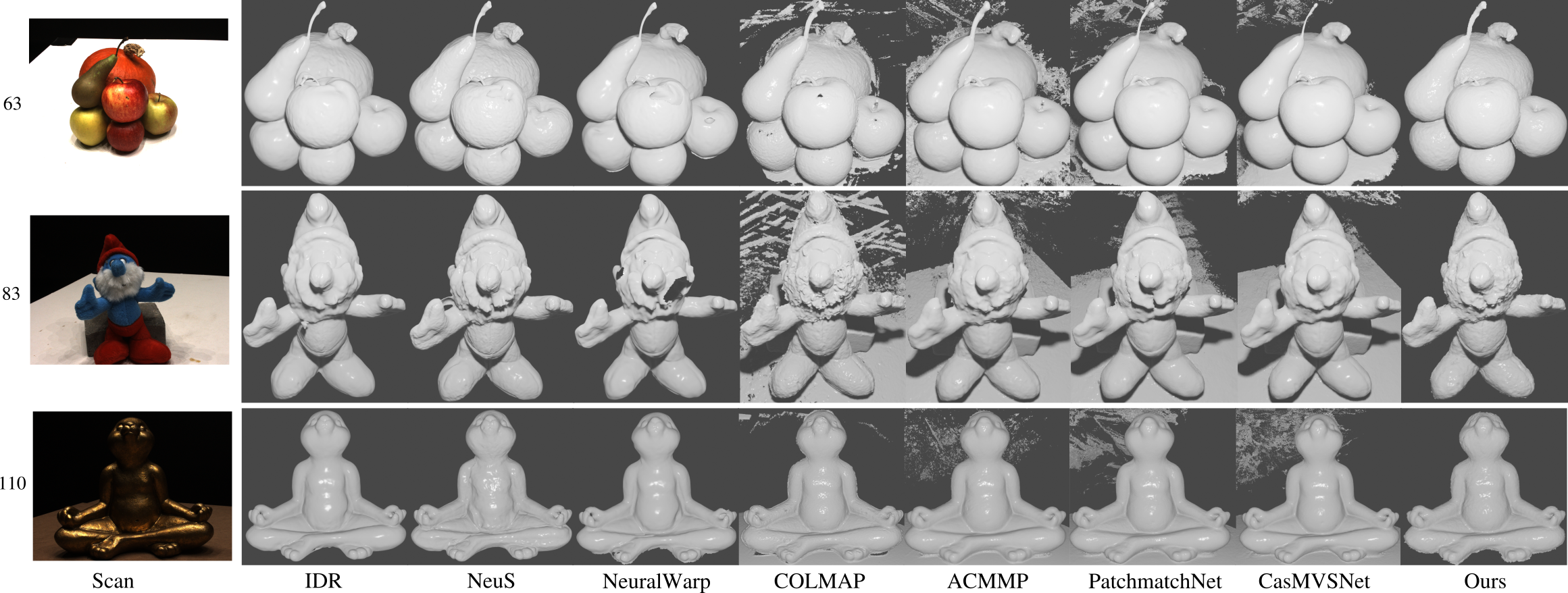}
\end{center}
\vspace{-3mm}
\caption{Qualitative comparisons with state-of-the-art methods.}
\label{fig:main}
\vspace{-3mm}
\end{figure*}

\subsection{Baseline Methods}
To assess our approach, we evaluate the geometry against state-of-the-art methods among 3 categories: classic MVS, deep MVS and differential rendering based methods. First, COLMAP~\cite{schonberger2016pixelwise} and ACMMP~\cite{xu2019multi} are classic MVS methods that have been widely used and demonstrate strong performances for MVS reconstruction. Among all the deep MVS methods, we consider two of the most efficient methods PatchmatchNet~\cite{wang2021patchmatchnet} and CasMVSNet~\cite{gu2020cascade} for which the code is available and easy to use. Finally, for the differentiable rendering based methods we consider IDR~\cite{yariv2020multiview} which was one of the first works that combines a differentiable surface renderer with a neural implicit representation. It requires accurate masks but handles specular surfaces and has shown impressive reconstruction results. We also compare with two more recent works that use volumetric rendering and provide impressive reconstruction results: NeuS~\cite{wang2021neus} and NeuralWarp~\cite{darmon2022improving}.

For the evaluation on DTU using all the available views ($49$ or $64$ depending on the scan) we retrain PatchmatchNet and CasMVSNet as their pre-trained models use a different train/test split. We use the pre-trained models for IDR, NeuS (with the mask loss) and NeuralWarp.

To recover meshes with our method, we use a post-processing step with a bilateral filter on the optimized depth maps, a TSDF Fusion~\cite{curless1996volumetric} method and a mesh cleaning based on the input masks. For COLMAP, ACMMP, PatchmatchNet and CasMVSNet we try to use the same TSDF Fusion method~\cite{curless1996volumetric} as much as possible. For differentiable rendering based methods (IDR, NeuS and NeuralWarp), the implicit representation is simply evaluated in a 3D grid of size $512^3$ and then Marching Cubes~\cite{lorensen-1987} is applied.

\subsection{Multi-view Reconstruction from Real Data}

\begin{figure}[!]
\begin{center}
\includegraphics[width=\linewidth]{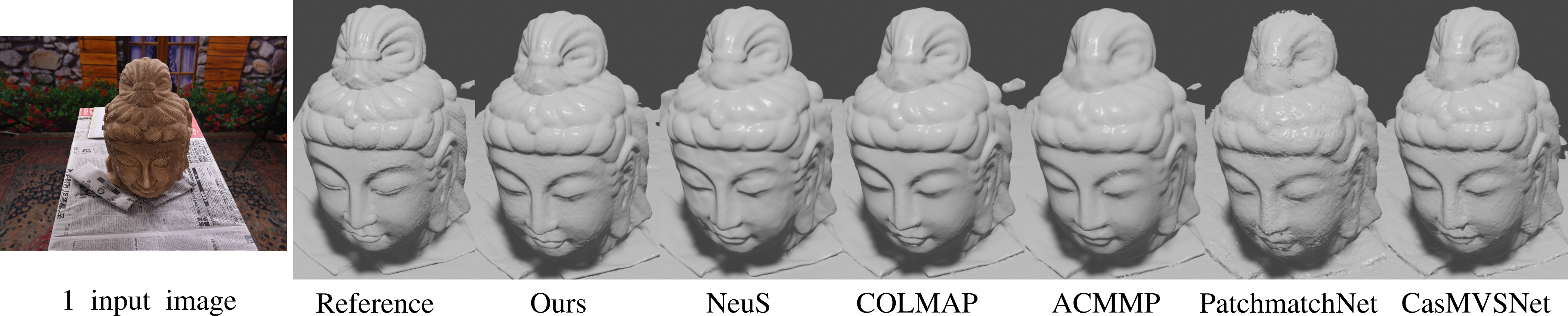}
\end{center}
\vspace{-3mm}
\caption{Qualitative comparison using 36 images of a model from BlendedMVS~\cite{yao2020blendedmvs}.}
\label{fig:blendedmvs}
\vspace{-2mm}
\end{figure}

\textbf{Qualitative results}. In Figure~\ref{fig:main}, we show comparisons between our method and the considered baselines. 
While IDR, NeuS and NeuralWarp produce high quality details, they show some artifacts on some misleading parts: some regions of the fruits (1st row), near the right arm of the figurine (2nd row) or at the separation between the belly and the legs of the statue (3rd row). In contrast, our method provides high level of details without failing on these difficult parts. For IDR and NeuS, the appearance prediction probably compensates for the wrong geometry during the optimization, however our approach exhibits more robustness by focusing on the geometry.
Our method produces visual results comparable to COLMAP, ACMMP, PatchmatchNet and CasMVSNet with high fidelity and reduced noise.
As shown in Figure~\ref{fig:teaser} and~\ref{fig:blendedmvs} our method provides similar results with high quality details on BlendedMVS, using the same photoconsistency prior trained on DTU. Note that in Figure~\ref{fig:teaser}, we use a coarse initial reconstruction obtained with~\cite{wang2021patchmatchnet}.

\textbf{Quantitative results}. 
In our quantitative evaluation, all the results are computed on the meshes obtained with the differentiable rendering based methods and directly on the point clouds fused from the depthmaps for COLMAP, ACMMP, PatchmatchNet, CasMVSNet and our method.
On average, our method clearly outperforms methods based on differentiable rendering (IDR, NeuS and NeuralWarp) in terms of accuracy and completeness. Our method also demonstrates an improvement over classic MVS methods COLMAP and ACMMP. Compared to deep MVS methods that are trained end-to-end on DTU, the approach is on par, though better on combined  accuracy and completeness, while training only a small neural network for photo-consistency that is used as a prior in the optimization. Note that quantitative results for PatchmatchNet and CasMVSNet are not as high as in their paper since the training set is not the same and in contrast to their train/test split, we remove all the scans from the training set in which a test object is seen. A detailed table showing the metrics computed for each scan is available in the supplementary.

\textbf{Runtime analysis}. In terms of runtime, our method (single machine and GPU) is competitive with the MVS methods COLMAP and ACMMP and takes around 45 minutes to optimize with $49$ images from DTU. We note that it could also benefit from an easy parallelization by optimizing different groups of depthmaps on different GPUs or even machines which can significantly reduce the computation time. COLMAP, ACMMP and our method have a strong computation time advantage compared to methods based on differentiable rendering and neural implicit representations (IDR, NeuS and NeuralWarp) that require several hours (between 10h and 25h) before converging to an accurate 3D reconstruction. Of course, deep MVS methods like PatchmatchNet and CasMVSNet that perform only inference are inherently much faster (a few seconds or minutes) than optimization based methods.

\begin{table}
\centering
\resizebox{0.7\linewidth}{!}{
\begin{small}
\begin{tabular}{|l||c||c c|}
    \hline
    Methods & Chamfer Distance $\downarrow$ & Accuracy $\downarrow$ & Completeness $\downarrow$\\
    \hline
IDR~\cite{yariv2020multiview} & 0.89 & 1.02 & 0.79 \\
NeuS~\cite{wang2021neus} & 0.77 & 0.85 & 0.68 \\
NeuralWarp~\cite{darmon2022improving} & 0.69 & 0.68 & 0.69 \\
COLMAP~\cite{schonberger2016pixelwise} & 0.49 & 0.40 & 0.58 \\
ACMMP~\cite{xu2019multi} &  0.42 & 0.46 & 0.39 \\
PatchmatchNet~\cite{wang2021patchmatchnet} & 0.40 & 0.44 & \textbf{0.35}\\
CasMVSNet~\cite{gu2020cascade} & \underline{0.38} & \textbf{0.34} & 0.43 \\
Ours & \textbf{0.36} & \underline{0.37} & \underline{0.36}\\
    \hline
\end{tabular}\end{small}
}

\caption{Quantitative evaluation on DTU~\cite{jensen2014large} ($49$ or $64$ images per model). Best scores are in \textbf{bold} and second best are \underline{underlined}.}
\vspace{-3mm}
\label{tab:eval_dtu}
\end{table}

\subsection{Reconstruction from Synthetic Data}

\begin{figure}[!]
\begin{center}
\includegraphics[width=\linewidth]{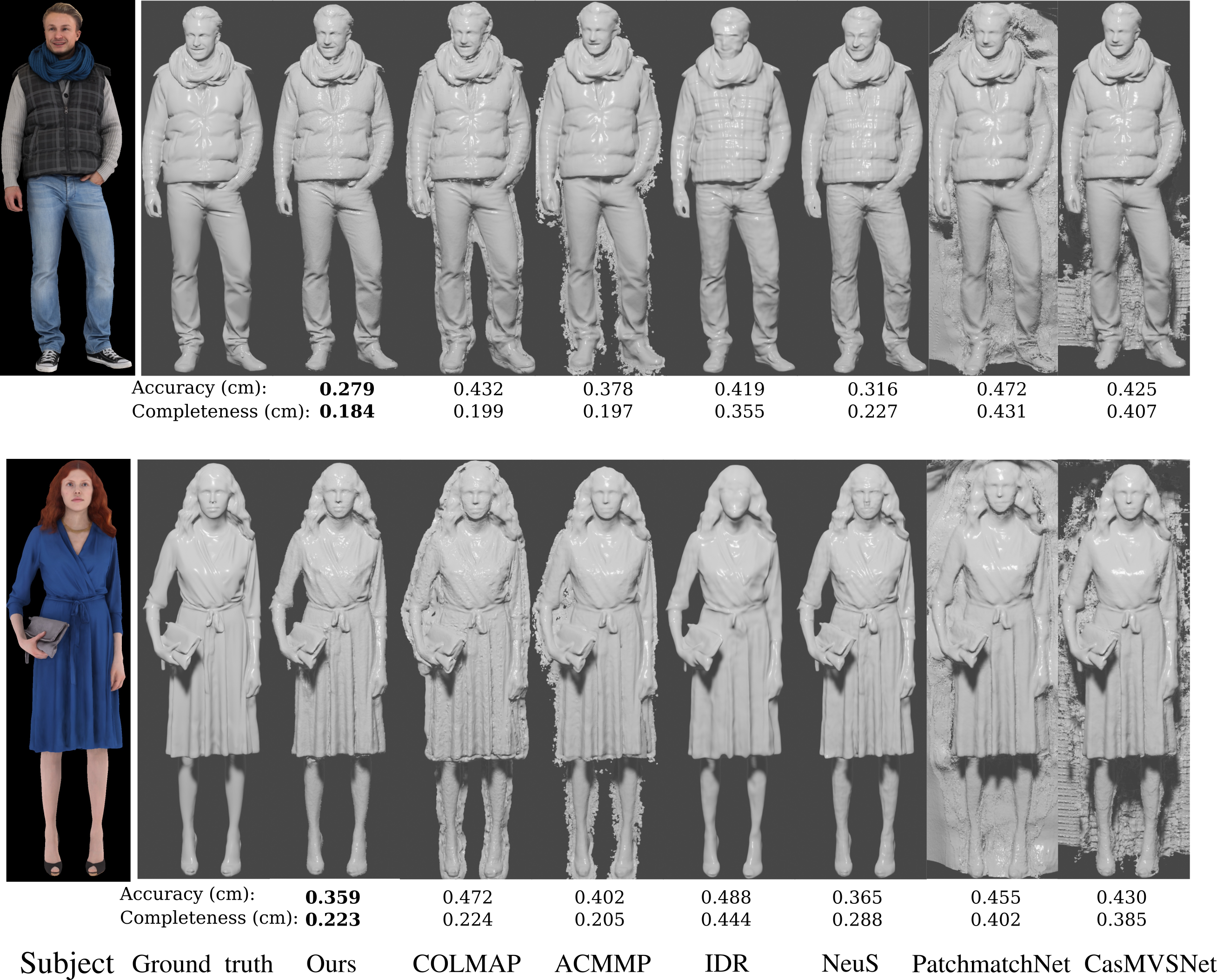}
\end{center}
\vspace{-3mm}
\caption{Qualitative and quantitative results with $19$ images (Renderpeople~\cite{renderpeople}) and using the baseline photo-consistency prior defined in~\ref{sec:photometric_prior}.}
\label{fig:rp}
\vspace{-2mm}
\end{figure}

To reinforce the validity of our volumetric shape energy, we experiment with the proposed baseline photo-consistency prior defined in Section~\ref{sec:photometric_prior}. 
We use $19$ synthetic images from Renderpeople~\cite{renderpeople} and compare qualitatively with classic MVS methods (COLMAP, ACMMP), differentiable rendering based methods (IDR, NeuS) and deep MVS methods (PatchmatchNet, CasMVSNet).

As shown in Figure~\ref{fig:rp}, our method is able to reconstruct very accurate and detailed meshes. COLMAP and ACMMP's are less detailed and more noisy (\eg COLMAP's bottom row). IDR and NeuS also lack details and even fail to reconstruct correctly the geometry of the jacket on the first row because of the checkered texture. In that case, optimizing both the geometry and the color clearly leads to the wrong geometry. PatchmatchNet and CasMVSNet also work well with very little noise (\eg feet on the first row) and slighlty less pronounced details compared to our method (\eg wrinkles on the scarf and sweater on the first row and on the upper part of the dress on the bottom row). The qualitative results are confirmed by the accuracy and completeness metrics computed between each reconstruction and the ground truth mesh.

\subsection{Reconstruction from Real Captured Data}
\label{sec:kinovis}

\begin{figure*}[h!]
\begin{center}
\includegraphics[width=0.95\linewidth]{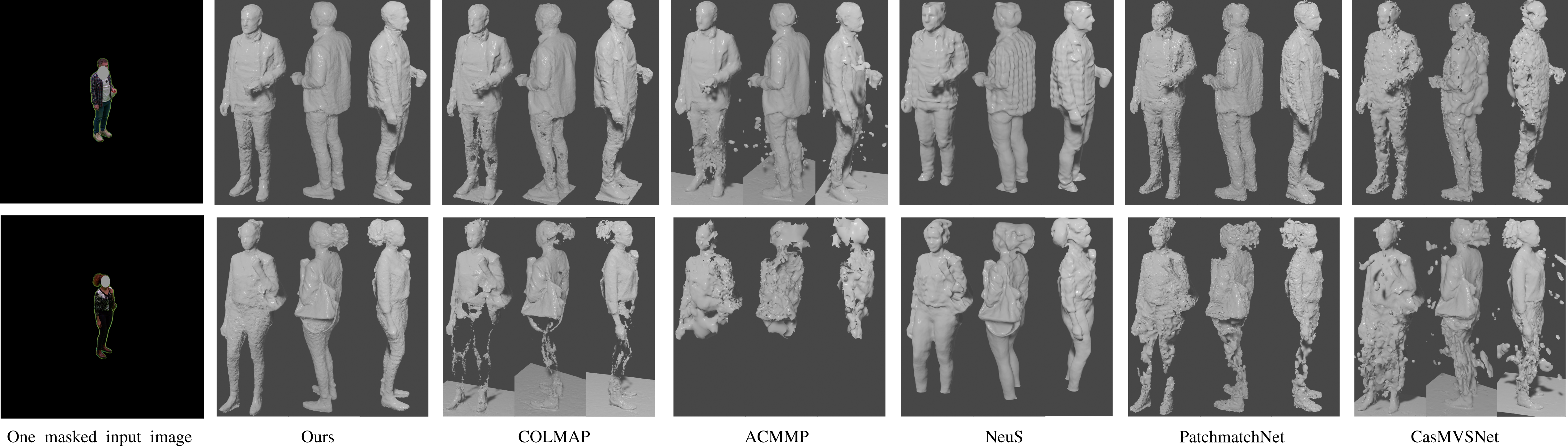}
\end{center}
\vspace{-4mm}
\caption{Left: One example image. Right: Qualitative comparison using $65$ images from a multi-camera platform.}
\label{fig:kinovis}
\vspace{-4mm}
\end{figure*}

To further evaluate the generalization ability of our method, we apply it on human capture data. We use images from a hemispherical multi-camera platform composed of $65$ cameras of various focal lengths. This setup is designed to capture humans moving in a large scene so the setting is significantly different from DTU with more distant cameras and significantly wider baselines.

Similarly to the previous experiments we compare with different methods: COLMAP, ACMMP, PatchmatchNet and CasMVSNet. For time reason, we only compare with one optimization method based on differentiable rendering. We choose Neus as it performs better than IDR on DTU and is much faster than NeuralWarp which requires an expensive two stage optimization. Note that PatchmatchNet, CasMVSNet and our learned photo-consistency prior are all trained on the same training set of DTU.

Qualitatively, COLMAP performs well with the top row model, despite some holes in the legs. However it has difficulties with the black pants and the hair with the bottom row model. ACMMP is less precise but we mention that a single optimization iteration was used due to RAM's limitation, even with 64Gb. NeuS reconstructs a nice watertight surface but lacks high-frequency details (\eg faces on both rows) and exhibits poor geometries at different locations due to  appearance ambiguities. The deep MVS methods PatchmatchNet and CasMVSNet partially suceed with the top example but fail with the bottom one. This illustrates the generalization issue with the full end-to-end learning based methods when the inference scenario is substantially different from the training one (\ie DTU). 
On the other hand, our method shows detailed surfaces with limited noise even on some difficult parts as the black pants on the bottom row. It demonstrates the benefit of a weaker prior with local photoconsistency, that is anyway embedded in a global optimization framework.

\subsection{Finetuning inference-based results}
\label{sec:finetuning}
\begin{figure}[h!]
\begin{center}
\includegraphics[width=0.7\linewidth]{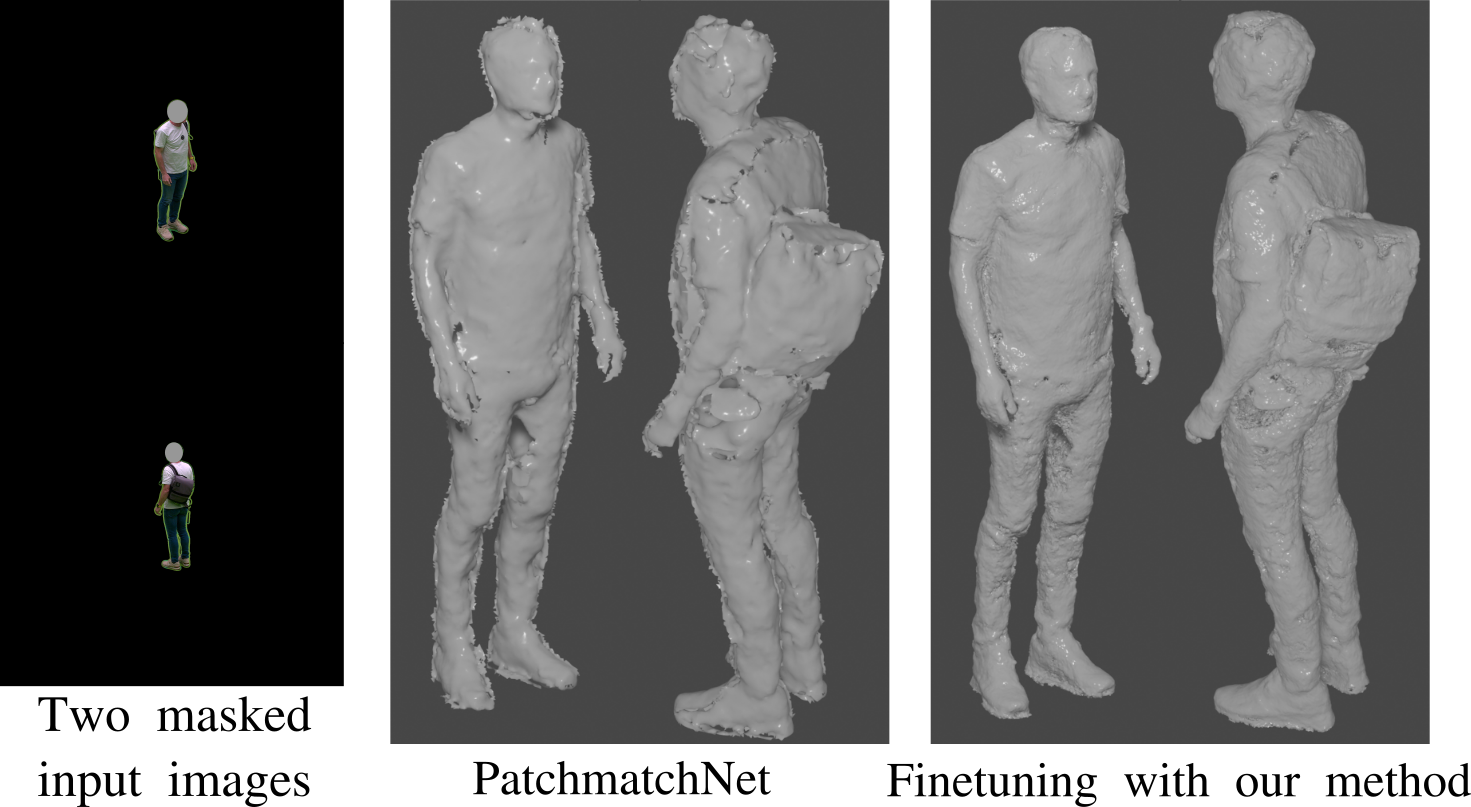}
\end{center}
\vspace{-3mm}
\caption{Finetuning the reconstruction result of PatchmatchNet with our optimization method.}
\label{fig:finetune}
\vspace{-2mm}
\end{figure}

Deep MVS methods like PatchmatchNet and CasMVSNet present the advantage of very fast inference but, as shown in the previous experiment~\ref{sec:kinovis}, tend to poorly generalize. From this observation, we experiment in this section the combination of an inference-based method with our optimization-based method. As shown in Fig.~\ref{fig:finetune}, the result of PatchmatchNet can be used as the initialization for the optimization rather than a coarse Visual Hull. Finetuning the results from PatchmatchNet exhibits more details and less noise but it fails to recover from large errors as with the top of the head, the top of the back and the left hip.

\section{Conclusion}
We have presented a strategy that combines depth optimization, as performed in the latest MVS strategies, with volumetric representations as used in more recent methods based on differentiable rendering. Building on signed distances our SRDF representation allows to optimize multi-view depthmaps in a consistent way by correlating depth prediction with photometric observations along viewing rays. Experiments on real and synthetic data demonstrate the efficiency of our method compared to classic MVS, deep MVS and differentiable rendering based methods. We also demonstrate the good applicability of our method with a learned photo-consistency prior that generalize well on data completely  different from the training set. As future work, we believe that photo-consistency priors can still be explored to improve generalization even further (data augmentation for example) or efficiency (specific architecture for very fast inference). We do not see any immediate negative societal impact of our method, but we still need to be very cautious as accurate 3D models of humans could be used maliciously (deep fakes, identity theft, ...).

\section{Acknowledgements}
We thank Laurence Boissieux and Julien Pansiot from the Kinovis platform at Inria Grenoble and our volunteer subjects for help with the 3D data acquisition.

{\small
\bibliographystyle{ieee_fullname}
\bibliography{egbib}

\begin{thebibliography}{10}\itemsep=-1pt

\bibitem{dtu_eval}
Dtueval-python.
\newblock \url{https://github.com/jzhangbs/DTUeval-python}.

\bibitem{renderpeople}
Renderpeople, 2018.
\newblock \url{https://renderpeople.com/3d-people/}.

\bibitem{agrawal2001probabilistic}
Motilal Agrawal and Larry~S Davis.
\newblock A probabilistic framework for surface reconstruction from multiple
  images.
\newblock In {\em IEEE Conference on Computer Vision and Pattern Recognition},
  volume~2, pages II--II. IEEE, 2001.

\bibitem{arandjelovic2021nerf}
Relja Arandjelovi{\'c} and Andrew Zisserman.
\newblock Nerf in detail: Learning to sample for view synthesis.
\newblock {\em arXiv preprint arXiv:2106.05264}, 2021.

\bibitem{broadhurst2001probabilistic}
Adrian Broadhurst, Tom~W Drummond, and Roberto Cipolla.
\newblock A probabilistic framework for space carving.
\newblock In {\em IEEE International Conference on Computer Vision}, volume~1,
  pages 388--393. IEEE, 2001.

\bibitem{campbell2008using}
Neill~DF Campbell, George Vogiatzis, Carlos Hern{\'a}ndez, and Roberto Cipolla.
\newblock Using multiple hypotheses to improve depth-maps for multi-view
  stereo.
\newblock In {\em European Conference on Computer Vision}, pages 766--779.
  Springer, 2008.

\bibitem{curless1996volumetric}
Brian Curless and Marc Levoy.
\newblock A volumetric method for building complex models from range images.
\newblock In {\em Annual Conference on Computer Graphics and Interactive
  Techniques}, pages 303--312, 1996.

\bibitem{darmon2022improving}
Fran{\c{c}}ois Darmon, B{\'e}n{\'e}dicte Bascle, Jean-Cl{\'e}ment Devaux,
  Pascal Monasse, and Mathieu Aubry.
\newblock Improving neural implicit surfaces geometry with patch warping.
\newblock In {\em Proceedings of the IEEE/CVF Conference on Computer Vision and
  Pattern Recognition}, pages 6260--6269, 2022.

\bibitem{de1999poxels}
Jeremy~S De~Bonet and Paul Viola.
\newblock Poxels: Probabilistic voxelized volume reconstruction.
\newblock In {\em International Conference on Computer Vision (ICCV)}, pages
  418--425, 1999.

\bibitem{dellaert2020neural}
Frank Dellaert and Lin Yen-Chen.
\newblock Neural volume rendering: Nerf and beyond.
\newblock {\em arXiv preprint arXiv:2101.05204}, 2020.

\bibitem{donne2019learning}
Simon Donne and Andreas Geiger.
\newblock Learning non-volumetric depth fusion using successive reprojections.
\newblock In {\em IEEE/CVF Conference on Computer Vision and Pattern
  Recognition}, pages 7634--7643, 2019.

\bibitem{furukawa2009accurate}
Yasutaka Furukawa and Jean Ponce.
\newblock Accurate, dense, and robust multiview stereopsis.
\newblock {\em IEEE Transactions on Pattern Analysis and Machine Intelligence},
  32(8):1362--1376, 2009.

\bibitem{gadelha20173d}
Matheus Gadelha, Subhransu Maji, and Rui Wang.
\newblock 3d shape induction from 2d views of multiple objects.
\newblock In {\em International Conference on 3D Vision}, pages 402--411. IEEE,
  2017.

\bibitem{galliani2016gipuma}
Silvano Galliani, Katrin Lasinger, and Konrad Schindler.
\newblock Gipuma: Massively parallel multi-view stereo reconstruction.
\newblock {\em Publikationen der Deutschen Gesellschaft f{\"u}r
  Photogrammetrie, Fernerkundung und Geoinformation e. V}, 25(361-369):2, 2016.

\bibitem{genova2019learning}
Kyle Genova, Forrester Cole, Daniel Vlasic, Aaron Sarna, William~T Freeman, and
  Thomas Funkhouser.
\newblock Learning shape templates with structured implicit functions.
\newblock In {\em IEEE/CVF International Conference on Computer Vision}, pages
  7154--7164, 2019.

\bibitem{gu2020cascade}
Xiaodong Gu, Zhiwen Fan, Siyu Zhu, Zuozhuo Dai, Feitong Tan, and Ping Tan.
\newblock Cascade cost volume for high-resolution multi-view stereo and stereo
  matching.
\newblock In {\em Proceedings of the IEEE/CVF Conference on Computer Vision and
  Pattern Recognition}, pages 2495--2504, 2020.

\bibitem{hartmann2017learned}
Wilfried Hartmann, Silvano Galliani, Michal Havlena, Luc Van~Gool, and Konrad
  Schindler.
\newblock Learned multi-patch similarity.
\newblock In {\em IEEE International Conference on Computer Vision}, pages
  1586--1594, 2017.

\bibitem{henderson2018learning}
Paul Henderson and Vittorio Ferrari.
\newblock Learning to generate and reconstruct 3d meshes with only 2d
  supervision.
\newblock {\em arXiv preprint arXiv:1807.09259}, 2018.

\bibitem{huang2018deepmvs}
Po-Han Huang, Kevin Matzen, Johannes Kopf, Narendra Ahuja, and Jia-Bin Huang.
\newblock Deepmvs: Learning multi-view stereopsis.
\newblock In {\em IEEE Conference on Computer Vision and Pattern Recognition},
  pages 2821--2830, 2018.

\bibitem{insafutdinov2018unsupervised}
Eldar Insafutdinov and Alexey Dosovitskiy.
\newblock Unsupervised learning of shape and pose with differentiable point
  clouds.
\newblock {\em Advances in Neural Information Processing Systems}, 31, 2018.

\bibitem{jang2021codenerf}
Wonbong Jang and Lourdes Agapito.
\newblock Codenerf: Disentangled neural radiance fields for object categories.
\newblock In {\em IEEE/CVF International Conference on Computer Vision}, pages
  12949--12958, 2021.

\bibitem{jensen2014large}
Rasmus Jensen, Anders Dahl, George Vogiatzis, Engin Tola, and Henrik Aan{\ae}s.
\newblock Large scale multi-view stereopsis evaluation.
\newblock In {\em IEEE Conference on Computer Vision and Pattern Recognition},
  pages 406--413, 2014.

\bibitem{jiang2018gal}
Li Jiang, Shaoshuai Shi, Xiaojuan Qi, and Jiaya Jia.
\newblock Gal: Geometric adversarial loss for single-view 3d-object
  reconstruction.
\newblock In {\em European Conference on Computer Vision}, pages 802--816,
  2018.

\bibitem{jimenez2016unsupervised}
Danilo Jimenez~Rezende, SM Eslami, Shakir Mohamed, Peter Battaglia, Max
  Jaderberg, and Nicolas Heess.
\newblock Unsupervised learning of 3d structure from images.
\newblock {\em Advances in Neural Information Processing Systems}, 29, 2016.

\bibitem{kato2018neural}
Hiroharu Kato, Yoshitaka Ushiku, and Tatsuya Harada.
\newblock Neural 3d mesh renderer.
\newblock In {\em IEEE Conference on Computer Vision and Pattern Recognition},
  pages 3907--3916, 2018.

\bibitem{kazhdan2006poisson}
Michael Kazhdan, Matthew Bolitho, and Hugues Hoppe.
\newblock Poisson surface reconstruction.
\newblock In {\em Eurographics Symposium on Geometry Processing}, volume~7,
  2006.

\bibitem{kazhdan2013screened}
Michael Kazhdan and Hugues Hoppe.
\newblock Screened poisson surface reconstruction.
\newblock {\em ACM Transactions on Graphics (ToG)}, 32(3):1--13, 2013.

\bibitem{kellnhofer2021neural}
Petr Kellnhofer, Lars~C Jebe, Andrew Jones, Ryan Spicer, Kari Pulli, and Gordon
  Wetzstein.
\newblock Neural lumigraph rendering.
\newblock In {\em IEEE/CVF Conference on Computer Vision and Pattern
  Recognition}, pages 4287--4297, 2021.

\bibitem{kutulakos2000theory}
Kiriakos~N Kutulakos and Steven~M Seitz.
\newblock A theory of shape by space carving.
\newblock {\em International Journal of Computer Vision}, 38(3):199--218, 2000.

\bibitem{leroy2018shape}
Vincent Leroy, Jean-S{\'e}bastien Franco, and Edmond Boyer.
\newblock Shape reconstruction using volume sweeping and learned
  photoconsistency.
\newblock In {\em European Conference on Computer Vision}, pages 781--796,
  2018.

\bibitem{li2021neural}
Tianye Li, Mira Slavcheva, Michael Zollhoefer, Simon Green, Christoph Lassner,
  Changil Kim, Tanner Schmidt, Steven Lovegrove, Michael Goesele, and Zhaoyang
  Lv.
\newblock Neural 3d video synthesis.
\newblock {\em arXiv preprint arXiv:2103.02597}, 2021.

\bibitem{liu2019soft}
Shichen Liu, Tianye Li, Weikai Chen, and Hao Li.
\newblock Soft rasterizer: A differentiable renderer for image-based 3d
  reasoning.
\newblock In {\em IEEE/CVF International Conference on Computer Vision}, pages
  7708--7717, 2019.

\bibitem{liu2020dist}
Shaohui Liu, Yinda Zhang, Songyou Peng, Boxin Shi, Marc Pollefeys, and Zhaopeng
  Cui.
\newblock Dist: Rendering deep implicit signed distance function with
  differentiable sphere tracing.
\newblock In {\em IEEE/CVF Conference on Computer Vision and Pattern
  Recognition}, pages 2019--2028, 2020.

\bibitem{lorensen-1987}
William~E. Lorensen and Harvey~E. Cline.
\newblock Marching cubes: A high resolution 3d surface construction algorithm.
\newblock In {\em SIGGRAPH}, pages 163--169, 1987.

\bibitem{merrell2007real}
Paul Merrell, Amir Akbarzadeh, Liang Wang, Philippos Mordohai, Jan-Michael
  Frahm, Ruigang Yang, David Nist{\'e}r, and Marc Pollefeys.
\newblock Real-time visibility-based fusion of depth maps.
\newblock In {\em IEEE International Conference on Computer Vision}, pages
  1--8. IEEE, 2007.

\bibitem{mescheder2019occupancy}
Lars Mescheder, Michael Oechsle, Michael Niemeyer, Sebastian Nowozin, and
  Andreas Geiger.
\newblock Occupancy networks: Learning 3d reconstruction in function space.
\newblock In {\em IEEE/CVF Conference on Computer Vision and Pattern
  Recognition}, pages 4460--4470, 2019.

\bibitem{michalkiewicz2019implicit}
Mateusz Michalkiewicz, Jhony~K Pontes, Dominic Jack, Mahsa Baktashmotlagh, and
  Anders Eriksson.
\newblock Implicit surface representations as layers in neural networks.
\newblock In {\em IEEE/CVF International Conference on Computer Vision}, pages
  4743--4752, 2019.

\bibitem{mildenhall2020nerf}
Ben Mildenhall, Pratul~P Srinivasan, Matthew Tancik, Jonathan~T Barron, Ravi
  Ramamoorthi, and Ren Ng.
\newblock Nerf: Representing scenes as neural radiance fields for view
  synthesis.
\newblock In {\em European Conference on Computer Vision}, pages 405--421.
  Springer, 2020.

\bibitem{muller2022instant}
Thomas M{\"u}ller, Alex Evans, Christoph Schied, and Alexander Keller.
\newblock Instant neural graphics primitives with a multiresolution hash
  encoding.
\newblock {\em arXiv preprint arXiv:2201.05989}, 2022.

\bibitem{murez2020atlas}
Zak Murez, Tarrence~van As, James Bartolozzi, Ayan Sinha, Vijay Badrinarayanan,
  and Andrew Rabinovich.
\newblock Atlas: End-to-end 3d scene reconstruction from posed images.
\newblock In {\em European Conference on Computer Vision}, pages 414--431.
  Springer, 2020.

\bibitem{navaneet2019capnet}
KL Navaneet, Priyanka Mandikal, Mayank Agarwal, and R~Venkatesh Babu.
\newblock Capnet: Continuous approximation projection for 3d point cloud
  reconstruction using 2d supervision.
\newblock In {\em AAAI Conference on Artificial Intelligence}, volume~33, pages
  8819--8826, 2019.

\bibitem{nguyen2018rendernet}
Thu~H Nguyen-Phuoc, Chuan Li, Stephen Balaban, and Yongliang Yang.
\newblock Rendernet: A deep convolutional network for differentiable rendering
  from 3d shapes.
\newblock {\em Advances in Neural Information Processing Systems}, 31, 2018.

\bibitem{DBLP:journals/corr/abs-1912-07372}
Michael Niemeyer, Lars~M. Mescheder, Michael Oechsle, and Andreas Geiger.
\newblock Differentiable volumetric rendering: Learning implicit 3d
  representations without 3d supervision.
\newblock {\em CoRR}, abs/1912.07372, 2019.

\bibitem{oechsle2019texture}
Michael Oechsle, Lars Mescheder, Michael Niemeyer, Thilo Strauss, and Andreas
  Geiger.
\newblock Texture fields: Learning texture representations in function space.
\newblock In {\em IEEE/CVF International Conference on Computer Vision}, pages
  4531--4540, 2019.

\bibitem{oechsle2021unisurf}
Michael Oechsle, Songyou Peng, and Andreas Geiger.
\newblock Unisurf: Unifying neural implicit surfaces and radiance fields for
  multi-view reconstruction.
\newblock In {\em IEEE/CVF International Conference on Computer Vision}, pages
  5589--5599, 2021.

\bibitem{park2019deepsdf}
Jeong~Joon Park, Peter Florence, Julian Straub, Richard Newcombe, and Steven
  Lovegrove.
\newblock Deepsdf: Learning continuous signed distance functions for shape
  representation.
\newblock In {\em IEEE/CVF Conference on Computer Vision and Pattern
  Recognition}, pages 165--174, 2019.

\bibitem{passalis2022opendr}
N Passalis, S Pedrazzi, R Babuska, W Burgard, D Dias, F Ferro, M Gabbouj, O
  Green, A Iosifidis, E Kayacan, et~al.
\newblock Opendr: An open toolkit for enabling high performance, low footprint
  deep learning for robotics.
\newblock {\em arXiv preprint arXiv:2203.00403}, 2022.

\bibitem{paszke2017automatic}
Adam Paszke, Sam Gross, Soumith Chintala, Gregory Chanan, Edward Yang, Zachary
  DeVito, Zeming Lin, Alban Desmaison, Luca Antiga, and Adam Lerer.
\newblock Automatic differentiation in pytorch.
\newblock 2017.

\bibitem{peng2020convolutional}
Songyou Peng, Michael Niemeyer, Lars Mescheder, Marc Pollefeys, and Andreas
  Geiger.
\newblock Convolutional occupancy networks.
\newblock In {\em European Conference on Computer Vision}, pages 523--540.
  Springer, 2020.

\bibitem{pumarola2021d}
Albert Pumarola, Enric Corona, Gerard Pons-Moll, and Francesc Moreno-Noguer.
\newblock D-nerf: Neural radiance fields for dynamic scenes.
\newblock In {\em IEEE/CVF Conference on Computer Vision and Pattern
  Recognition}, pages 10318--10327, 2021.

\bibitem{reiser2021kilonerf}
Christian Reiser, Songyou Peng, Yiyi Liao, and Andreas Geiger.
\newblock Kilonerf: Speeding up neural radiance fields with thousands of tiny
  mlps.
\newblock In {\em IEEE/CVF International Conference on Computer Vision}, pages
  14335--14345, 2021.

\bibitem{rich20213dvnet}
Alexander Rich, Noah Stier, Pradeep Sen, and Tobias H{\"o}llerer.
\newblock 3dvnet: Multi-view depth prediction and volumetric refinement.
\newblock In {\em International Conference on 3D Vision}, pages 700--709. IEEE,
  2021.

\bibitem{riegler2017octnetfusion}
Gernot Riegler, Ali~Osman Ulusoy, Horst Bischof, and Andreas Geiger.
\newblock Octnetfusion: Learning depth fusion from data.
\newblock In {\em International Conference on 3D Vision}, pages 57--66. IEEE,
  2017.

\bibitem{saito2019pifu}
Shunsuke Saito, Zeng Huang, Ryota Natsume, Shigeo Morishima, Angjoo Kanazawa,
  and Hao Li.
\newblock Pifu: Pixel-aligned implicit function for high-resolution clothed
  human digitization.
\newblock In {\em IEEE/CVF International Conference on Computer Vision}, pages
  2304--2314, 2019.

\bibitem{saito2020pifuhd}
Shunsuke Saito, Tomas Simon, Jason Saragih, and Hanbyul Joo.
\newblock Pifuhd: Multi-level pixel-aligned implicit function for
  high-resolution 3d human digitization.
\newblock In {\em IEEE/CVF Conference on Computer Vision and Pattern
  Recognition}, pages 84--93, 2020.

\bibitem{schonberger2016pixelwise}
Johannes~L Sch{\"o}nberger, Enliang Zheng, Jan-Michael Frahm, and Marc
  Pollefeys.
\newblock Pixelwise view selection for unstructured multi-view stereo.
\newblock In {\em European Conference on Computer Vision}, pages 501--518.
  Springer, 2016.

\bibitem{seitz2006}
Steven~M. Seitz, Brian Curless, James Diebel, Daniel Scharstein, and Richard
  Szeliski.
\newblock A comparison and evaluation of multi-view stereo reconstruction
  algorithms.
\newblock {\em IEEE Conference on Computer Vision and Pattern Recognition},
  1:519--528, 2006.

\bibitem{seitz1999photorealistic}
Steven~M Seitz and Charles~R Dyer.
\newblock Photorealistic scene reconstruction by voxel coloring.
\newblock {\em International Journal of Computer Vision}, 35(2):151--173, 1999.

\bibitem{sitzmann2019scene}
Vincent Sitzmann, Michael Zollh{\"o}fer, and Gordon Wetzstein.
\newblock Scene representation networks: Continuous 3d-structure-aware neural
  scene representations.
\newblock {\em Advances in Neural Information Processing Systems}, 32, 2019.

\bibitem{sun2021neuralrecon}
Jiaming Sun, Yiming Xie, Linghao Chen, Xiaowei Zhou, and Hujun Bao.
\newblock Neuralrecon: Real-time coherent 3d reconstruction from monocular
  video.
\newblock In {\em IEEE/CVF Conference on Computer Vision and Pattern
  Recognition}, pages 15598--15607, 2021.

\bibitem{takikawa2021neural}
Towaki Takikawa, Joey Litalien, Kangxue Yin, Karsten Kreis, Charles Loop, Derek
  Nowrouzezahrai, Alec Jacobson, Morgan McGuire, and Sanja Fidler.
\newblock Neural geometric level of detail: Real-time rendering with implicit
  3d shapes.
\newblock In {\em IEEE/CVF Conference on Computer Vision and Pattern
  Recognition}, pages 11358--11367, 2021.

\bibitem{tulsiani2017multi}
Shubham Tulsiani, Tinghui Zhou, Alexei~A Efros, and Jitendra Malik.
\newblock Multi-view supervision for single-view reconstruction via
  differentiable ray consistency.
\newblock In {\em IEEE Conference on Computer Vision and Pattern Recognition},
  pages 2626--2634, 2017.

\bibitem{wang2021patchmatchnet}
Fangjinhua Wang, Silvano Galliani, Christoph Vogel, Pablo Speciale, and Marc
  Pollefeys.
\newblock Patchmatchnet: Learned multi-view patchmatch stereo.
\newblock In {\em Proceedings of the IEEE/CVF Conference on Computer Vision and
  Pattern Recognition}, pages 14194--14203, 2021.

\bibitem{wang2021neus}
Peng Wang, Lingjie Liu, Yuan Liu, Christian Theobalt, Taku Komura, and Wenping
  Wang.
\newblock Neus: Learning neural implicit surfaces by volume rendering for
  multi-view reconstruction.
\newblock {\em arXiv preprint arXiv:2106.10689}, 2021.

\bibitem{xu2019multi}
Qingshan Xu and Wenbing Tao.
\newblock Multi-scale geometric consistency guided multi-view stereo.
\newblock In {\em Proceedings of the IEEE/CVF Conference on Computer Vision and
  Pattern Recognition}, pages 5483--5492, 2019.

\bibitem{xu2019disn}
Qiangeng Xu, Weiyue Wang, Duygu Ceylan, Radomir Mech, and Ulrich Neumann.
\newblock Disn: Deep implicit surface network for high-quality single-view 3d
  reconstruction.
\newblock {\em Advances in Neural Information Processing Systems}, 32, 2019.

\bibitem{yao2018mvsnet}
Yao Yao, Zixin Luo, Shiwei Li, Tian Fang, and Long Quan.
\newblock Mvsnet: Depth inference for unstructured multi-view stereo.
\newblock In {\em European Conference on Computer Vision (ECCV)}, pages
  767--783, 2018.

\bibitem{yao2019recurrent}
Yao Yao, Zixin Luo, Shiwei Li, Tianwei Shen, Tian Fang, and Long Quan.
\newblock Recurrent mvsnet for high-resolution multi-view stereo depth
  inference.
\newblock In {\em IEEE/CVF Conference on Computer Vision and Pattern
  Recognition}, pages 5525--5534, 2019.

\bibitem{yao2020blendedmvs}
Yao Yao, Zixin Luo, Shiwei Li, Jingyang Zhang, Yufan Ren, Lei Zhou, Tian Fang,
  and Long Quan.
\newblock Blendedmvs: A large-scale dataset for generalized multi-view stereo
  networks.
\newblock In {\em Proceedings of the IEEE/CVF Conference on Computer Vision and
  Pattern Recognition}, pages 1790--1799, 2020.

\bibitem{yariv2021volume}
Lior Yariv, Jiatao Gu, Yoni Kasten, and Yaron Lipman.
\newblock Volume rendering of neural implicit surfaces.
\newblock volume~34, 2021.

\bibitem{yariv2020multiview}
Lior Yariv, Yoni Kasten, Dror Moran, Meirav Galun, Matan Atzmon, Basri Ronen,
  and Yaron Lipman.
\newblock Multiview neural surface reconstruction by disentangling geometry and
  appearance.
\newblock In {\em Advances in Neural Information Processing Systems},
  volume~33, 2020.

\bibitem{yu2021plenoxels}
Alex Yu, Sara Fridovich-Keil, Matthew Tancik, Qinhong Chen, Benjamin Recht, and
  Angjoo Kanazawa.
\newblock Plenoxels: Radiance fields without neural networks.
\newblock {\em arXiv preprint arXiv:2112.05131}, 2021.

\bibitem{yu2021plenoctrees}
Alex Yu, Ruilong Li, Matthew Tancik, Hao Li, Ren Ng, and Angjoo Kanazawa.
\newblock Plenoctrees for real-time rendering of neural radiance fields.
\newblock In {\em IEEE/CVF International Conference on Computer Vision}, pages
  5752--5761, 2021.

\bibitem{yu2021pixelnerf}
Alex Yu, Vickie Ye, Matthew Tancik, and Angjoo Kanazawa.
\newblock pixelnerf: Neural radiance fields from one or few images.
\newblock In {\em IEEE/CVF Conference on Computer Vision and Pattern
  Recognition}, pages 4578--4587, 2021.

\bibitem{zagoruyko2015learning}
Sergey Zagoruyko and Nikos Komodakis.
\newblock Learning to compare image patches via convolutional neural networks.
\newblock In {\em IEEE Conference on Computer Vision and Pattern Recognition},
  pages 4353--4361, 2015.

\bibitem{zhu2018visual}
Jun-Yan Zhu, Zhoutong Zhang, Chengkai Zhang, Jiajun Wu, Antonio Torralba, Josh
  Tenenbaum, and Bill Freeman.
\newblock Visual object networks: Image generation with disentangled 3d
  representations.
\newblock {\em Advances in Neural Information Processing Systems}, 31, 2018.

\bibitem{zobeidi2021deep}
Ehsan Zobeidi and Nikolay Atanasov.
\newblock A deep signed directional distance function for object shape
  representation.
\newblock {\em arXiv preprint arXiv:2107.11024}, 2021.

\end{thebibliography}
}

\end{document}